%% file: main.tex
\documentclass[10pt,journal,compsoc]{IEEEtran}
\usepackage{amsmath}
\usepackage{amssymb}
\usepackage{algorithm}
\usepackage{algorithmic}
\usepackage[group-separator={,}]{siunitx}         

\usepackage{graphicx}
\usepackage{subcaption}
\usepackage{threeparttable}

\usepackage{amsmath}
\usepackage{amsfonts}
\usepackage{amsthm}
\usepackage{multirow}
\usepackage{bm}
\usepackage{bbm}
\usepackage{diagbox}

\usepackage{booktabs}

\newcommand{\E}{\mathbb{E}}
\newcommand{\R}{\mathbb{R}}
\def \xx {\mathbf{x}}
\def \xa {\mathbf{x}_{adv}}

\def \ya {y_{adv}}
\def \zz {\mathbf{z}}
\def \bb {\mathbf{b}}

\def \gg {\mathbf{g}}
\def \hh {\mathbf{h}}

\def \ss {\mathbf{s}}
\def \uu {\mathbf{u}}
\def \vv {\mathbf{v}}

\def \X  {\mathcal{X}}
\def \Y  {\mathcal{Y}}

\def \R  {\mathbb{R}}

\def \rr  {\mathbf{r}}
\def \N  {\mathcal{N}}

\def \bU  {\mathbf{U}}

\def \S  {\mathcal{S}}

\def \la {\ell_{adv}}
\def \trans {\mathsf{T}}
\DeclareMathOperator*{\argmin}{arg\,min}

\DeclareMathOperator{\sign}{sign}
\newcommand{\norm}[1]{\left\lVert#1\right\rVert}
\usepackage{xspace}
\def\eg{\emph{e.g.}\xspace} 

\def\ie{\emph{i.e.}\xspace}

\begin{document}
\title{Black-box Adversarial Attacks on Video Recognition Models}

\author{Linxi Jiang$^{*1}$, Xingjun Ma$^{*2}$, Shaoxiang Chen$^{1}$, James Bailey$^{2}$, Yu-Gang Jiang$^{\dag1}$

\thanks{$^\star$ indicates equal contributions.}
\thanks{$^\dag$ indicates corresponding author.}

\IEEEcompsocitemizethanks{\IEEEcompsocthanksitem Linxi Jiang, Shaoxiang Chen, and Yu-Gang Jiang are with the School of Computer Science, Fudan University, Shanghai, China.
\protect\\ E-mail: \{lxjiang18, sxchen13, ygj\}@fudan.edu.cn.
\IEEEcompsocthanksitem Xingjun Ma, James Bailey are with The University of Melbourne. \protect\\ Email: \{xingjun.ma, baileyj\}@unimelb.edu.au.}
}

\IEEEtitleabstractindextext{%
\begin{abstract}
Deep neural networks (DNNs) are known for their vulnerability to adversarial examples. These are examples that have undergone small, carefully crafted perturbations, and which can easily fool a DNN into making misclassifications at test time. 
Thus far, the field of adversarial research has mainly focused on image models, under either a white-box setting, where an adversary has full access to model parameters, or a black-box setting where an adversary can only query the target model for probabilities or labels. Whilst several white-box attacks have been proposed for video models,  black-box video attacks are still unexplored. To close this gap, we propose the first black-box video attack framework, called V-BAD.
V-BAD utilizes \textit{tentative perturbations} transferred from image models, and \textit{partition-based rectifications} found by the NES on partitions (patches) of tentative perturbations, to obtain good adversarial gradient estimates with fewer queries to the target model. V-BAD is equivalent to estimating the projection of an adversarial gradient on a selected subspace. Using three benchmark video datasets, we demonstrate that V-BAD can craft both untargeted and targeted attacks to fool two state-of-the-art deep video recognition models. For the targeted attack, it achieves  $>$93\% success rate using only an average of $3.4 \sim 8.4 \times 10^4$ queries, a similar number of queries to state-of-the-art black-box image attacks. This is despite the fact that videos often have two orders of magnitude higher dimensionality than static images. We believe that V-BAD is a promising new tool to evaluate and improve the robustness of video recognition models to black-box adversarial attacks.
\end{abstract}
}

\maketitle

\IEEEdisplaynontitleabstractindextext

\IEEEpeerreviewmaketitle

\IEEEraisesectionheading{\section{Introduction}
\label{sec:intro}}
\IEEEPARstart{D}{eep} Neural Networks (DNNs) are a family of powerful models that have demonstrated superior performance in a wide range of visual understanding tasks that has been extensively studied in both the multimedia and computer vision communities such as video recognition\cite{karpathy2014large,ji20133d,carreira2017quo,wu2016multi}, image classification\cite{krizhevsky2012imagenet,zhao2017learning} and video captioning\cite{yang2017catching,liu2018sibnet}. 
Despite their current success, DNNs have been found to be extremely vulnerable to adversarial examples (or attacks) \cite{szegedy2013intriguing,goodfellow2014explaining}. For DNN classifiers, adversarial examples can be easily generated by applying adversarial perturbations to clean (normal) samples, that maximize the classification error \cite{goodfellow2014explaining,nguyen2015deep,carlini2017towards}. For images, the perturbations are often small and visually imperceptible to human observers, but they can fool DNNs into making misclassifications with high confidence. The vulnerability of DNNs to adversarial examples has raised serious security concerns for their deployment in security-critical applications, such as face recognition \cite{kurakin2016adversarial} and self-driving cars\cite{eykholt2017robust}. Hence, the study of adversarial examples for DNNs has become a crucial task for secure deep learning.

Adversarial examples can be generated by an attack method (also called an adversary) following either a white-box setting (white-box attacks) or a black-box setting (black-box attacks). In the white-box setting, an adversary has full access to the target model (the model to attack), including model parameters and training settings.   In the black-box setting, an adversary only has partial information about the target model, such as the labels or probabilities output by the model. White-box methods generate an adversarial example by applying one step or multiple steps perturbations on a clean test sample, following the direction of the adversarial gradient \cite{goodfellow2014explaining,carlini2017towards}. The adversarial gradient is the gradient of an adversarial loss, which is typically defined to maximize (rather than minimize) classification error. However, in the black-box setting, adversarial gradients are not accessible to an adversary. In this case, the adversary can first attack a local surrogate model and then transfer these attacks to  the target model \cite{moosavi2017universal,papernot2016transferability,papernot2017practical}.  Alternatively they may use a black-box optimization method such as Finite Differences (FD) or Natural Evolution Strategies (NES), to estimate the gradient \cite{bhagoji2018practical,ilyas2018black,chen2017zoo}. 

A number of attack methods have been proposed \cite{nguyen2015deep,carlini2017towards,madry2017towards}, however, most of them focus on either image models, or video models but in a white-box setting \cite{li2018adversarial,rey2018targeted}. Different from these works, in this paper, we propose a framework for the generation of adversarial attacks against video recognition models, specifically in a black-box setting.
Significant progress has been achieved for black-box image attacks, but not for black-box video attacks.  A key reason is that videos typically have much higher dimensionality (often two magnitudes higher) than static images. On static images, for most attacks to succeed, existing black-box methods must use $\sim 10^4$ queries \cite{bhagoji2018practical,chen2017zoo} on CIFAR-10 \cite{krizhevsky2009learning} images, and $\sim 10^5$ queries \cite{ilyas2018black} on ImageNet \cite{deng2009imagenet} images. Due to their massive input dimensions, black-box attacks on videos generally require two orders of magnitude more queries for gradient estimation than are needed for images. This makes black-box video attacks impractical, taking into account time and budget constraints. To better evaluate the robustness of video models, it is therefore important to explore efficient black-box methods that can generate attacks using fewer queries.

\begin{figure}
\centering
  \includegraphics[width=0.9\linewidth]{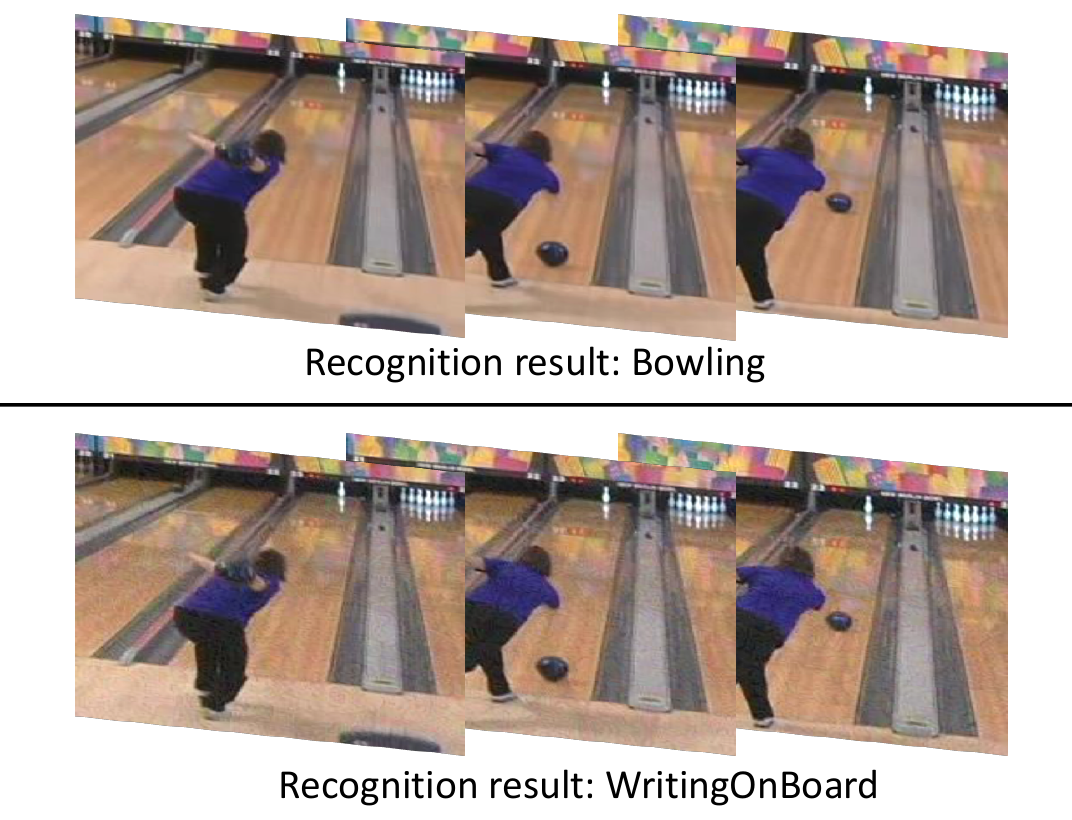}
  \caption{An example of black-box video adversarial attacks (targeted). The original video (top) can be correctly recognized while the adversarial one (bottom) generated by our proposed method is misclassified by the same video model.}
  \label{fig:advimgs}
\end{figure}

\begin{figure*}
  \includegraphics[width=\textwidth]{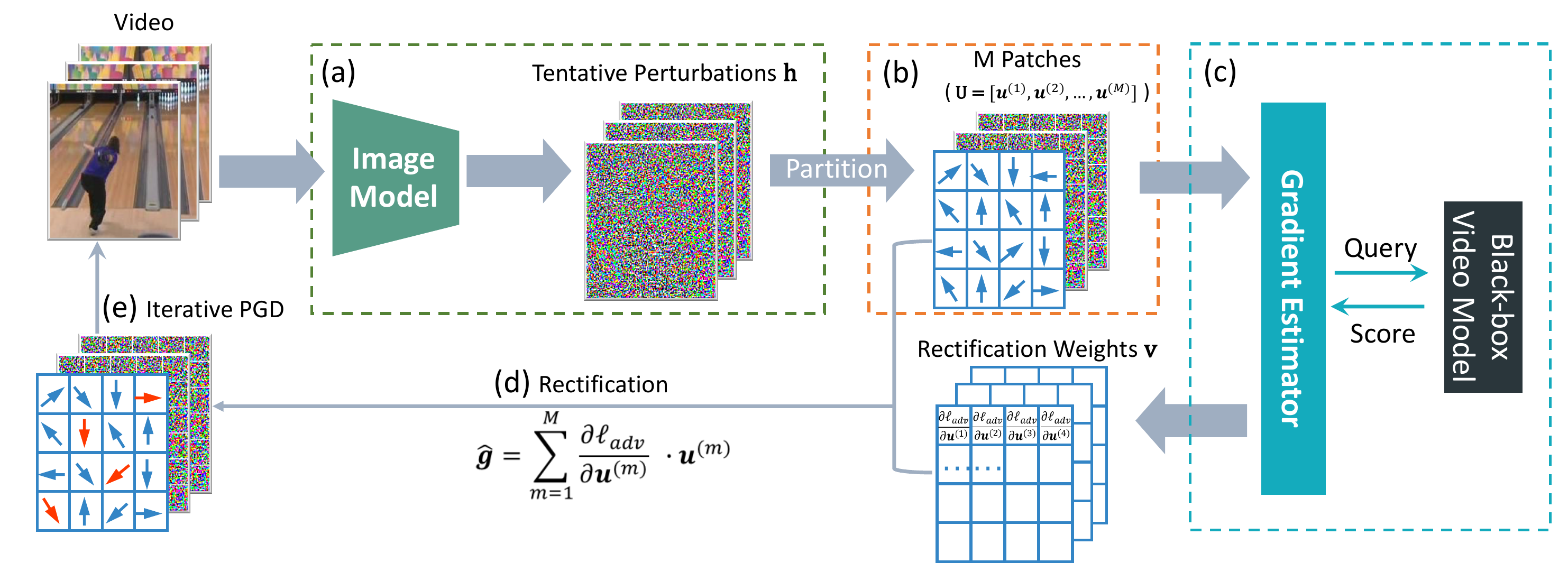}
  \caption{Overview of the proposed V-BAD framework for black-box  video attacks.}
  \label{fig:framework}
\end{figure*}

In this paper, we propose a simple and efficient framework for the generation of black-box adversarial attacks on video recognition models. Intuitively, we exploit the transferability of adversarial perturbations and the derivative-free optimization methods to obtain accurate estimations of the adversarial gradients. In particular, we first generate tentative perturbations as a rough estimate of the true adversarial gradient using ImageNet-pretrained DNNs.  We then rectify these tentative perturbations in \emph{patches} (or partitions) using NES, by querying the target model. Our proposed framework only needs to estimate a small number of directional derivatives (of the patches) rather than estimating pixel-wise derivatives, making it an efficient framework for black-box video attacks. Figure \ref{fig:advimgs} shows an example of video adversarial attacks generated by our proposed method. In summary, our main contributions are:
\begin{itemize}
 \item We study the problem of black-box attacks on video recognition models and propose a general framework called V-BAD, to generate black-box video adversarial examples. To the best of our knowledge, our proposed framework V-BAD is the first black-box adversarial attack framework for videos. 

 \item Our proposed framework V-BAD exploits both the transferability of adversarial perturbations via the use of tentative perturbations, and the advantages of gradient estimation via NES for patch-level rectification of the tentative perturbations.
 We also show that V-BAD is equivalent to estimating the projection of the adversarial gradient on a selected subspace.

 \item We conduct an empirical evaluation using three benchmark video datasets and two state-of-the-art video recognition models.  We show that V-BAD can achieve high attack success rates with few queries to target models, making it a useful tool for the robustness evaluation of video models.
\end{itemize}

\section{Related Work}

\textbf{White-box Image Attack.}
The fast gradient sign method (FGSM) crafts an adversarial example by perturbing a normal sample along the gradient direction towards maximizing the classification error \cite{goodfellow2014explaining}.
FGSM is a single-step attack, and can be applied iteratively to improve adversarial strength \cite{kurakin2016adversarial}. Projected Gradient Descent (PGD) \cite{madry2017towards} is another iterative method that is regarded as the strongest first-order attack. PGD projects the perturbations back onto the $\epsilon$-ball centered at the original sample when perturbations go beyond the $\epsilon$-ball. The C\&W attack solves the attack problem via an optimization framework \cite{carlini2017towards}, and is arguably the state-of-the-art white-box attack. There also exists other types of white-box methods, \eg, Jacobian-based Saliency Map Attack (JSMA) \cite{papernot2016limitations}, DeepFool \cite{moosavi2016deepfool} and elastic-net attack (EAD) \cite{chen2018ead}.

\textbf{Black-box Image Attack.}
In the black-box setting, the adversarial gradients are not directly accessible by an adversary. As such, black-box image attacks either exploit the transferability of adversarial examples (also known as transferred attacks) or make use of gradient estimation techniques. It was first observed in \cite{szegedy2013intriguing} that adversarial examples are transferable across models, even if the models have different architectures or were trained separately. \cite{papernot2017practical} trains a surrogate model locally on synthesized data, with labels obtained by querying the target model. It then generates adversarial examples from the surrogate model using white-box methods to attack the target model. However, training a surrogate model on synthesized data often incurs a huge number of queries, and the transferability of generated adversarial examples is often limited. \cite{chen2017zoo} proposes the use of Finite Differences (FD), a black-box gradient estimation method, to estimate the adversarial gradient. \cite{bhagoji2018practical} accelerates FD-based gradient estimation with dimensionality reduction techniques such as PCA. Compared to FD, \cite{ilyas2018black} demonstrates improved performance with fewer queries by the use of the other type of gradient estimation method called Natural Evolutionary Strategies (NES).

\textbf{White-box Video Attack.}
In contrast to image adversarial examples, much less work has been done for video adversarial examples. White-box video attacks were first investigated in \cite{wei2018sparse}, which discussed the sparsity and propagation of adversarial perturbations across video frames. \cite{li2018adversarial} leverages Generative Adversarial Networks (GANs) to perturb each frame in real-time video classification. In this paper, we explore black-box attacking methods against state-of-the-art video recognition models, which, to the best of our knowledge, is the first work on black-box video attacks.

\textbf{Video Recognition Models.} 
Encouraged by the great success of Convolutional Neural Networks (CNNs) on image recognition tasks, many works propose to adapt image (2D) CNNs to video recognition. \cite{karpathy2014large} explores various approaches for extending 2D CNNs to video recognition based on features extracted from individual frames.
CNN+LSTM based models \cite{jiang2018exploiting,yue2015beyond} exploit the temporal information contained in successive frames, with recurrent layers capturing long term dependencies on top of CNNs. 
C3D\cite{ji20133d,tran2015learning,varol2018long} extends the 2D spatio-only filters in traditional CNNs to 3D spatio-temporal filters for videos, and learn hierarchical spatio-temporal representations directly from videos.
However, C3D models often have a huge number of parameters, which makes training difficult. To address this, \cite{carreira2017quo} proposes the Inflated 3D ConvNet(I3D) with Inflated 2D filters and pooling kernels of traditional 2D CNNs. In this paper, we use two representative state-of-the-art video recognition models, CNN+LSTM and I3D, as our target models to attack.

\section{Proposed Framework V-BAD}
\subsection{Preliminaries}
We denote a video sample by $\xx \in \X \subset \R^{N\times H \times W \times C}$ with $N$, $H$, $W$, $C$ denoting the number of frames, frame height, frame width, and the number of channels respectively, and its associated true class by $y \in \Y= \{1, \cdots, K\}$. Video recognition is to learn a DNN classifier $f(\xx; \mathbf{\theta}): \X \to \Y$ by minimizing the classification loss $\ell(f(\xx;\mathbf{\theta}), y)$, and $\mathbf{\theta}$ denotes the parameters of the network. When the context is clear, we abbreviate $f(\xx; \mathbf{\theta})$ as $f(\xx)$, and $\ell(f(\xx;\mathbf{\theta}), y)$ as $\ell(\xx, y)$. The goal of adversarial attack is to find an adversarial example $\xa$ that can fool the network to make a false prediction, while keeping the adversarial example $\xa$  within the $\epsilon_{adv}$-ball centered at the original example $\xx$ ($\|\xa - \xx\|_p \leq \epsilon_{adv}$). Following early works \cite{ilyas2018black,madry2017towards,lidadversarial,athalye2018obfuscated,wang2019convergence}, in this paper, we only focus on the $L_\infty$-norm, that is, $\|\xa - \xx\|_\infty \leq \epsilon_{adv}$, but our framework also applies to other norms. 

There are two types of adversarial attacks: untargeted attack and targeted attack. Untargeted attack is to find an adversarial example that can be misclassified as any class other than the correct one (\eg $f(\xa) \neq y$), while targeted attack is to find an adversarial example that can be misclassified as a targeted adversarial class (\eg $f(\xa) = \ya$ and $\ya \neq y$). For simplicity, we denote the adversarial loss function that should be optimized to find an adversarial example by $\la(\xx)$, and let $\la(\xx) = -\ell(\xx, y)$ for untargeted attack and $\la(\xx) = \ell(\xx, \ya)$ for targeted attack. We also denote the adversarial gradient of the adversarial loss to the input as $\gg=\nabla_{\xx}\la(\xx)$. Accordingly, an attacking method is to \emph{minimize} the adversarial loss $\la(\xx)$ by iteratively perturbing the input sample following the direction of the adversarial gradient $\gg$.

\textbf{Threat Model.} Our threat model follows the query-limited black-box setting as follows. The adversary takes the video classifier $f$ as a black-box oracle and only has access to its output of the top 1 score. More specifically, during the attack process, given an arbitrary clean sample $\xx$, the adversary can query the target model $f$ to obtain the top 1 label $\hat{y}$ and its probability $P(\hat{y}|\xx)$. The adversary is asked to generate attacks within $Q$ number of queries. We consider both untargeted and targeted black-box attacks. 

\subsection{Framework Overview}
The structure of the proposed framework, namely V-BAD, for black-box video attacks is illustrated in Figure \ref{fig:framework}.
Following steps (a)-(e) highlighted in the figure, V-BAD perturbs an input video iteratively as follow: (a) It passes video frames into a public image model (such as ImageNet-pretrained networks) to obtain pixel-wise \textit{tentative perturbations} $\hh$; (b) It then partitions pixel-wise tentative perturbations into a set of $M$ patches $\bU$ ($\bU=[\uu^{(1)}, \uu^{(2)}, ..., \uu^{(M)}]$  where $\uu^{(m)}$ represents the $m$-th patch); (c) A black-box gradient estimator estimates the weight (\eg $\hat{\vv}_m=\frac{\partial \la}{\partial \uu^{(m)}}$) to each patch for rectification (or correction), via querying the target video model; (d) The patch-wise rectification weights are applied on the patches to obtain the rectified pixel-wise perturbations $\hat{\gg}$; (e) Apply one step PGD update on the input video, according to the rectified perturbations $\hat{\gg}$.
Specifically, the proposed method at the $t$-th PGD perturbation step can be described as:
\begin{align}
    \vv &= \mathbf{0}\\
    \hh^t &= \phi(\xx^{t-1})\\
    \bU &= G(\hh^t) \\
    \hat{\vv} &= \vv + \nabla_{\vv}\la(\xx^{t-1}+R(\vv, \bU))\\
    \hat{\gg} &= R(\hat{\vv}, \bU) \\
    \xx^{t} &= \Pi_{\epsilon}(\xx^{t-1} - \alpha\cdot\sign(\hat{\gg}))
\end{align}
 where, $\xx^{t-1}$ is the adversarial example generated at step $t-1$, $\phi(\cdot)$ is the function to extract tentative perturbations (\eg $\hh^{t}$), $G(\cdot)$ is a partitioning method that splits pixels of $\hh^{t}$ into a set of patches $\bU$, $\hat{\vv}$ is the estimated rectification weights for patches in $\bU$ via a black-box gradient estimator, $R(\vv, \bU)$ is a rectification function that applies patch-wise rectifications $\hat{\vv}$ to patches in $\bU$ to obtain pixel-wise rectified perturbations (\eg $\hat{\gg}$), $\Pi(\cdot)$ is a projection operation, $\text{sign}(\cdot)$ is the sign function, $\alpha$ is the PGD step size, $\epsilon$ is the perturbaiton bound. $R(\vv, \bU)$ applies the rectification weights to patches, thus enables us to estimate the gradients with respect to the patches in place of the gradients with respect to the raw pixels, which reduces the dimension of attack space from $\R^{N\times H\times W\times C}$ to $\R^M$ where $M$ is the number of patches. Each of the above operations will be explained in the following sections.

\textbf{Targeted V-BAD Attack.}
For a targeted attack, we need to ensure that the target class remains in the top-1 classes during the attacking process, as the score of the target class is required for gradient estimation. Thus, instead of the original sample $\xx$, we begin with a sample from the target class (\eg $\xx^0 = \xx'$ and $f(\xx') = \ya$), then gradually (step by step) reduce the perturbations bound $\epsilon$ from $1$ (for normalized inputs $\xx \in [0, 1]$) to $\epsilon_{adv}$ while maintaining the targeted class as the top-1 class. Note that although we begin with $\xx^0 = \xx'$, the adversarial example is bounded within the $\epsilon_{adv}$-ball centered at the original example $\xx$. The targeted V-BAD attack is described in Algorithm \ref{alg:framework}. We use an epsilon decay $\Delta\epsilon$ to control the reduction size of the perturbation bound. The epsilon decay $\Delta\epsilon$ and PGD step size $\alpha$ are dynamically adjusted as described in \ref{sec:setting}.

\textbf{Untargeted V-BAD Attack.}
For untargeted attack, we use the original clean example as our starting point: $\xx^{0} = \xx$, where $\xx$ is the original clean example. And the perturbation bound $\epsilon$ is set to the constant $\epsilon_{adv}$ throughout the attacking process.

\input{pseudocode/framework}

\subsection{Tentative Perturbations}\label{sec:tent} Tentative perturbations refer to initialized pixel-wise adversarial perturbations, generated each time before rectification from image models. This corresponds to the function $\phi(\cdot)$ in Eq. (2). As PGD only needs the sign of adversarial gradient to perturb a sample, here we also take the sign value of the generated perturbations as the tentative perturbations.

Here, we discuss three types of tentative perturbations. 1) Random: the perturbation for each input dimension takes on the value 1 or -1 equiprobably. Random perturbations only provide a stochastic exploration of the input space, which can be extremely inefficient due to the massive input dimensions of videos. 2) Static: the perturbation for each input dimension is fixed to 1. Fixed perturbations impose a strong constraint on the exploration space, with the same tentative perturbations used for each gradient estimation.
3) Transferred: tentative perturbations can alternatively be transferred from off-the-shelf pre-trained image models.
Since natural images share certain similar patterns, image adversarial examples often transfer across models or domains, though the transferability can be limited and will vary depending on the content of the images. To better exploit such transferability for videos, we propose to white-box attack a pre-trained image model such as an ImageNet pre-trained DNN, in order to extract the tentative perturbations for each frame. 
The resulting transferred perturbations can provide useful guidance for the exploration space and help reduce the number of queries. 

The tentative perturbations for an intermediate perturbed sample $\xx^{t-1}$ can be extracted by:
\begin{equation}\small
  \phi(\xx^{t-1}) = \text{sign} \Big(\nabla_{\xx^{t-1}} \frac{1}{N}\sum_{n=1}^{N} \|\bb_n \circ \tilde{f}_{l}(\xx^{t-1}_n) - \bb_n \circ \rr_n\|_2^2 \Big),
\end{equation}
where $\text{sign}(\cdot)$ is the sign function, $\circ$ is the element-wise product, $\| \cdot \|_2^2$ is the squared $L_2$-norm, $N$ is the total number of frames, $\tilde{f}_{l}(\cdot)$ is the $l$-th layer output (\eg deep features) of a public image model $\tilde{f}$, $\bb_n$ is a random mask on the $n$-th frame, and $\rr_n$ is a target feature map which has the same dimensions as $\tilde{f}_{l}(\xx^{t-1}_n)$. 
$\phi(\cdot)$ is to generate tentative perturbations by minimizing the $L_2$ distance between the feature map $\tilde{f}_{l}(\cdot)$ and a target (or adversarial) feature map $\rr_n$.
For untargeted attack, $\rr_n$ is a feature map of Gaussian random noise, while for targeted attack, $\rr_n$ is the feature map $\tilde{f}_{l}(\xx'_n)$ of the $n$-th frame of a video $\xx'$ from the target class $\ya$. The random mask $\bb_n$ is a pixel-wise mask with each element is randomly set to either 1 or 0. The use of random mask can provide some randomness to the exploration of the attack space and help avoid getting stuck in situations where the frame-wise tentative perturbations are extremely inaccurate to allow any effective rectification. 

\subsection{Partition-based Rectification}\label{sec:group_rect}
Tentative perturbations are transferred rough estimate of the true adversarial gradient, thus it should be further rectified to better approximate the true adversarial gradient.
To achieve this more efficiently, we propose to perform rectifications at the patch-level (in contrast to pixel-level) and use gradient estimation methods to estimate the proper rectification weight for each patch.
That is, we first partition tentative perturbations into separate parts, which we call perturbation patches, and then estimate the rectification weights for these patches. In this subsection, we will introduce the partitioning methods, rectification function, and rectification weights estimation successively.

\textbf{Partitioning Methods.} We consider three types of partitioning strategies (\eg function $G(\cdot)$ in Eq. (3)). 1) \emph{Random}: dividing input dimensions randomly into a certain number of patches. Note that the input dimensions within one patch can be nonadjacent, although we refer to it as a patch. Random partitioning does not consider the local correlations between input dimensions but can be used as a baseline to assess the effectiveness of carefully designed partitioning methods. 2) \emph{Uniform}: splitting a frame uniformly into a certain number of patches. This will produce frame patches that preserve local dimensional correlations. 3) \emph{Semantic}: partitioning the video input according to its semantic content. It is promising to explore the correlation between the semantic content of the current input and its adversarial gradient.
In this paper, we empirically investigate the first two partitioning strategies, \ie, Random and Uniform, and leave semantic partitioning for future work.

\input{pseudocode/rectify}

Each patch constitutes a vector with the same number of dimensions as the input by zero padding: assigning zero values to dimensions that do not belong to the patch, while keeping its values at dimensions that belong to the patch. That is, for each patch, we have a vector $\uu^{(m)}$:
\begin{equation*}\label{eq:direction vector}
    \uu^{(m)}_j =
    \begin{cases}
        \hh_j & \text{if dimension $j$ is in the $m$-th partition} \\
        0 & \text{otherwise}.
    \end{cases}
\end{equation*}
Note that, $\uu^{(m)}$ has equal number of dimensions to both $\xx$ and $\hh$, namely, $\uu^{(m)} \in \R^{N \times H \times W \times C}$. We normalize the $\uu^{(m)}$ to unit vector $\uu^{(m)} = \frac{\uu^{(m)}}{|\uu^{(m)}|}$ ($|\cdot|$ is the vector norm), which we call the direction vector of the $m$-th patch.
The partition function can be written as:
\begin{equation*}
    G(\hh) = \big[\uu^{(1)}, \uu^{(2)}, ..., \uu^{(M)}\big].
\end{equation*}
We denote the output of partition function as $\bU = G(\hh)$.

\textbf{Rectification Function.}
The rectification function $R(\vv, \bU)$ (in Eq. (4) and (5)) can be expressed as applying the components of rectification vector $\vv$ as weights to each patch. 
Given a patch-wise rectification vector $\vv$ (estimated by some black-box gradient estimator), rectification is to apply patch weights to the direction vectors of the patches:
\begin{equation*}
    R(\vv, \bU) = \big[\uu^{(1)}, \uu^{(2)}, ..., \uu^{(M)}\big] \vv.
\end{equation*}

With the partition and rectification functions, we can optimize the rectification weights $\vv$ over patches instead of all input dimensions. This effectively reduces the dimensionality of the exploration space from $N\times H\times W\times C$ to $M$.

\textbf{Rectification Weights Estimation.}
For the estimation of rectification weights for patches, we propose to use the NES estimator which has been shown to be more efficient than FD in black-box image attacks \cite{ilyas2018black}. We will discuss the efficiency of FD compared to NES in Section \ref{sec:experiments}. Instead of maximizing the adversarial objective directly, NES maximizes the expected value of the objective under a search distribution. Different from \cite{ilyas2018black} where the gradient of adversarial loss with respect to the input is estimated, we estimate the gradient with respect to the patch weights $\vv$.
For a adversarial loss function $\la(\cdot)$, current parameters $\vv$, $\xx$, $\bU$ and a search distribution $\pi(\bm\gamma|\vv)$, we have:
\begin{equation*}
\begin{aligned}
     \nabla_{\vv} \E_{\pi(\bm\gamma|\vv)}&\big[\la(\xx + R(\bm\gamma, \bU) )\big] = \\ &\E_{\pi(\bm\gamma|\vv)}\big[\la(\xx + R(\bm\gamma, \bU) )\nabla_{\vv} \log\big(\pi(\bm\gamma|\vv)\big)\big].
\end{aligned}
\end{equation*}
Following \cite{ilyas2018black,salimans2017evolution,wierstra2014natural}, we use the normal distribution as the search distribution, that is, $\bm\gamma = \vv + \sigma\bm\delta$ where $\sigma$ is the search variance and $\bm\delta \sim \N(0, I)$ (standard normal distribution). We use antithetic sampling to generate a population of $\lambda$ number of $\bm\delta_k$ values: first sample Gaussian noise for $k \in \{1, \cdots, \frac{\lambda}{2}\}$, then set $\bm\delta_{k}=-\bm\delta_{\lambda-k+1}$ for $k \in \{(\frac{\lambda}{2} + 1), \cdots, \lambda\}$. 
Evaluating the gradient with a population of $\lambda$ points sampled under this scheme yields the following gradient estimate:
\begin{equation*}\small
    \nabla_{\vv} \E \big[\la(\xx + R(\bm\gamma, \bU) )\big] \approx \frac{1}{\lambda\sigma }\sum_{k=1}^\lambda \bm\delta_k \la(\xx + R(\vv + \sigma\bm\delta_{k}, \bU)).
\end{equation*}
Within each PGD perturbation step, $\vv$ is initialized to all zero values (see Eq. (1)) such that the starting point $\xx + R(\vv, \bU))$ is centered at $\xx$.

The complete estimation algorithm for patch rectification is described in Algorithm \ref{alg:nes}, where the function $\text{TransformedAdvLoss}(\cdot)$ is a ranking-based nonlinear transformation on the adversarial loss. The transformed loss increases monotonically with the original adversarial loss \cite{wierstra2014natural}, except for targeted attack, it also ``punishes'' $\bm\delta_{k}$ that fails to maintain the target class as the top-1 class with the highest loss value.
With this estimated gradient for $\vv$, we can update the rectification weights as $\hat{\vv} = \vv + \nabla_{\vv} \E [\la(\xx + R(\bm\gamma, \bU) )]$. We then use $\hat{\vv}$ to rectify tentative perturbations $\bU$ to $R(\hat{\vv}, \bU)$, which allows us to apply one step of PGD perturbation following Eq. (6).

\begin{figure}
  \centering
  \includegraphics[width=0.85\linewidth]{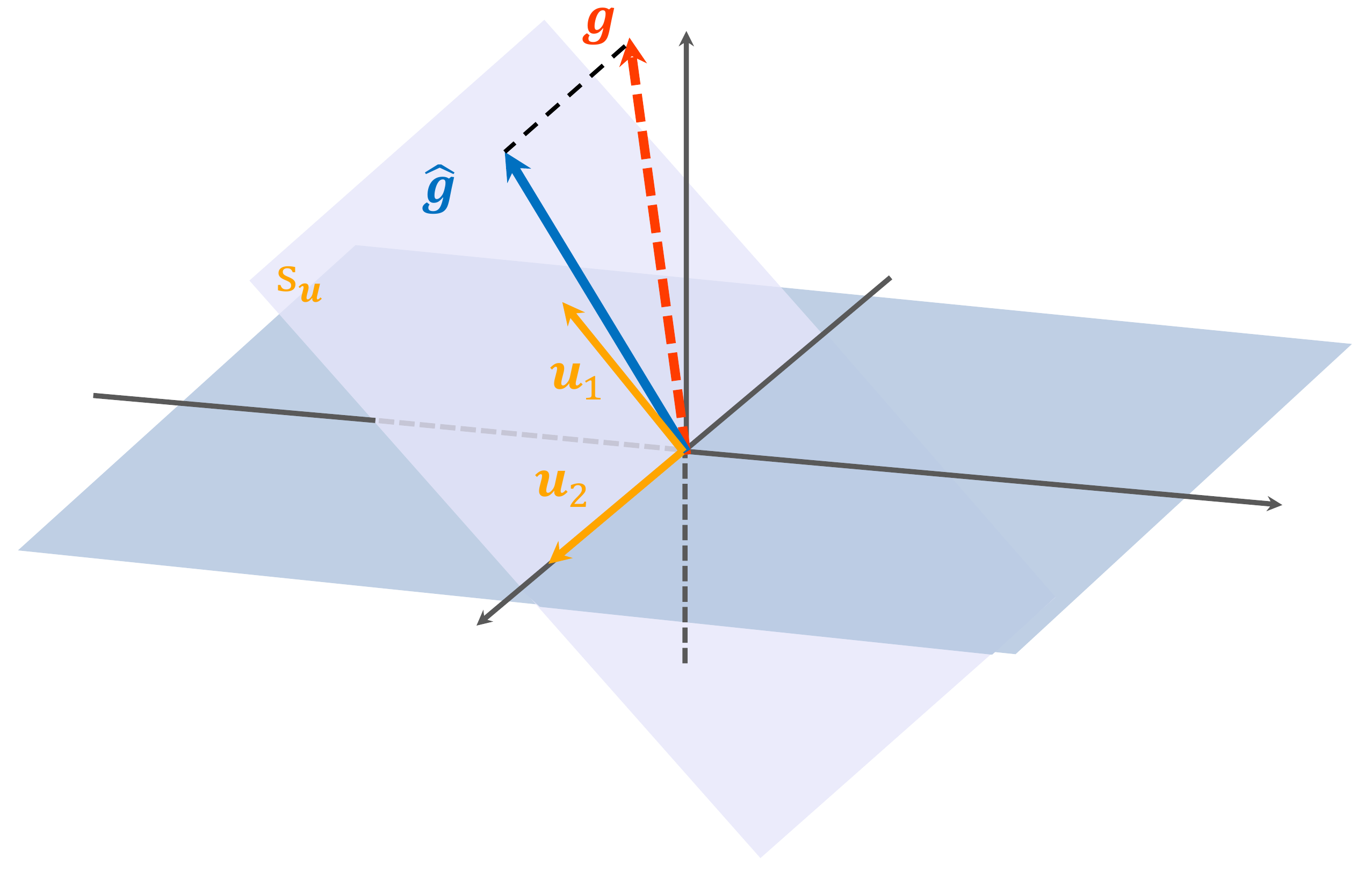}
  \caption{Projection ($\hat{\gg}$) of gradient $g$ on subspace $\ss_\uu$.}
  \label{fig:proj}
\end{figure}

\subsection{Analysis of V-BAD}\label{sec_analysis}
We prove that patch-rectified perturbations is the estimation of the projection of the adversarial gradient on a selected subspace. Let $\S$ be the input space with respect to $\X$, direction vectors (\eg $\uu$) of the tentative perturbation patches define a vector subspace $\ss_{\uu} \subseteq \S$. 
Using the chain rule, we can show that the gradients with respect to the patch-wise rectification weights $\vv$ are the directional derivatives of the patch directions $\uu$ evaluated at $\vv=\mathbf{0}$:
\begin{align*}
    \frac{\partial \la(\xx + R(\vv, \bU))}{\partial \vv_m}  = (\uu^{(m)})^{\trans} \nabla_{\xx} \la(\xx)
    = \frac{\partial \la(\xx)}{\partial \uu^{(m)}}.
\end{align*}
Suppose we have a perfect estimation of gradients with regards to $\vv$, then the rectified perturbations become:
\begin{align*}
    \hat{\gg} = R(\hat{\vv}, \bU) = \sum_{m=1}^{M} \uu^{(m)} \cdot \Big(\frac{\partial \la(\xx)}{\partial \uu^{(m)}}\Big).
\end{align*}
Since $\frac{\partial \la(\xx)}{\partial \uu^{(m)}}$ is the directional derivative of $\uu^{(m)}$, $\uu^{(m)} \cdot (\frac{\partial \la(\xx)}{\partial \uu^{(m)}})$ is the projection of adversarial gradient on the direction $\uu^{(m)}$, if the model is differentiable. 
Thus, rectified perturbations $\hat{\gg}$ is the vector sum of the gradient projection on each patch direction $\uu^{(m)}$. Therefore, rectified perturbation $\hat{\gg}$ can be regarded as the projection of adversarial gradient $\gg$ on subspace $\ss_{\uu}$, as the projection of a vector on a subspace is the vector sum of the vector's projections on the orthogonal basis vectors of the subspace.
Figure \ref{fig:proj} illustrates a toy example of the projection, where $\gg$ is the adversarial gradient, $\hat{\gg}$ is the projection of $\gg$ on a subspace $\ss_{\uu}$ that is defined by the direction vectors of two patches: $\uu^{(1)}, \uu^{(2)}$.

The main difference between our proposed patch-based rectification and the direct estimation of adversarial gradient over individual pixels is that patch rectification decouples the adversarial gradient estimation into two parts: 1) compressing the estimation space into subspaces $\ss_{\uu}$ defined by direction vectors of patches $\uu$, and 2) estimating the directional derivatives of direction vectors (\eg $\nabla_{\vv}\la(\xx+R(\vv, \bU))$). In an extreme case, each input dimension (or pixel) is a patch, the rectified perturbations becomes exactly the adversarial gradient, given perfect estimation of $\nabla_{\vv}\la(\xx+R(\vv, \bU))$. In a typical case, patch rectification is equivalent to estimating the projection of the adversarial gradient on a subspace $\ss_{\uu}$. A beneficial property of the projection $\hat{\gg}$ is that it is the closest vector to the gradient in subspace $\ss_{\uu}$ having $\hat{\gg} = \mathop{\argmin}_{\bm\eta}(\norm{\gg - \bm{\eta}}_2), \forall \bm{\eta} \in \ss_{\uu} \subseteq \S$. This enables us to only consider the subspace $\ss_{\uu}$ to find a good estimation of the adversarial gradient $\gg$.

\section{Experiments}\label{sec:experiments}
In this section, we provide a comprehensive evaluation of our proposed V-BAD framework and its variants, for both untargeted and targeted video attacks on three benchmark video datasets, against two state-of-the-art video recognition models. We also investigate different choices of tentative perturbations, partitioning methods and estimation methods in an ablation study.

\subsection{Experimental Setting}\label{sec:setting}
\textbf{Datasets.} We consider three benchmark datasets for video recognition: UCF-101 \cite{soomro2012ucf101}, HMDB-51 \cite{kuehne2011hmdb}, and Kinetics-400 \cite{kay2017kinetics}. UCF-101 is an action recognition dataset of \num{13320} realistic action videos, collected from YouTube. It consists of 101 action categories ranging from human-object/human- interaction, body-motion, playing musical instruments to sports. HMDB-51 is a dataset for human motion recognition, which contains \num{6849} clips from 51 action categories including facial actions and body movements, with each category containing a minimum of 101 clips. Kinetics-400 is also a dataset for human action recognition, which consists of approximately \num{240000} video clips from 400 human action classes with about 400 video clips (10 seconds) for each class. At test time, we use 32-frame snippets for UCF-101 and HMDB-51, and 64-frame snippets for Kinetics-400, as it has longer videos.

\textbf{Video Recognition Models.} We consider two state-of-the-art video recognition models I3D and CNN+LSTM, as our target models to attack. I3D is an inflated 3D convolutional network. We use a Kinetics-400 pretrained I3D and finetune it on other two datasets. We sample frames at 25 frames per second. CNN+LSTM is a combination of the conventional 2D convolutional network and the LSTM network. For CNN+LSTM we use a ImageNet pretrained ResNet101 as a frame feature extractor, then finetune a LSTM built on it. Input video frames are subsampled by keeping one out of every 5 for CNN+LSTM Model. Note we only consider the RGB part for both video models.
The test accuracy of the two models can be found in Table \ref{tab:model}. The accuracy gap between ours and the one reported in \cite{carreira2017quo} is mainly caused by the availability of fewer input frames at test time.

\textbf{Image Models.} We use ImageNet \cite{deng2009imagenet} pretrained deep networks as our image model for the generation of the tentative perturbations. ImageNet is an image dataset that contains more than 10 million natural images from more than 1000 classes. Given the difference between our video datasets and ImageNet, rather than using one model, we chose an ensemble of ImageNet models as our image model: ResNet50 \cite{he2016deep}, DenseNet121 \cite{huang2017densely}, and DenseNet169 \cite{huang2017densely}.

\textbf{Attack Setting.} For each dataset, we randomly select one test video, from each category, that is correctly classified by the target model. 
We randomly choose a target class for each video in targeted attack. 
For all datasets, we set the maximum adversarial perturbations magnitude to $\epsilon=0.05$ per frame. The query limit, \ie, the maximum number of queries to the target model is set to $Q=3\times10^5$, which is similar to the number of queries required for most black-box image attacks to succeed. We run the attack until an adversarial example is found (attack succeeds) or we reach the query limit. We evaluate different attack strategies in terms of 1) success rate (SR), the ratio of successful generation of adversarial examples under the perturbations bound within the limit of number of queries; and 2) average number of queries (ANQ), required for a successful attack (excluding failed attacks).

\begin{table}[t]
  \caption{Test Accuracy (\%) of the video models.}
  \label{tab:model}
  \centering
  \small
  \begin{tabular}{cccc}
    \toprule
    Model & UCF-101 & HMDB-51 & Kinetics-400\\
    \midrule
    I3D & 91.30 & 63.73 & 64.71 \\
    CNN+LSTM & 76.29 & 44.38 & 53.20\\
    \bottomrule
  \end{tabular}
\end{table}
\begin{table}
\centering
\small
  \caption{Results for V-BAD with different: 1) tentative perturbations, 2) partitioning methods, and 3) estimation methods. Symbol ``+" indicates fixed methods for untested components of V-BAD. Best results are highlighted in bold. ANQ: average number of queries; SR: success rate.}
  \label{tab:tent}
  \begin{tabular}{c|c|rr}
    \toprule
    \multirow{2}{*}{Components of V-BAD} & \multirow{2}{*}{Method} & \multicolumn{2}{c}{UCF-101} \\ 
    & & ANQ & SR (\%) \\ \hline \hline
    \multirow{4}{*}{\shortstack[l]{Tentative Perturbations\\+ partition: Uniform\\+ estimation: NES}} & Static & 51786 & 70 \\
    & Random & 107499 & 95  \\
    & Single & 57361 & 100 \\
    & \textbf{Ensemble} & \textbf{49797} & \textbf{100} \\ \hline \hline
    \multirow{3}{*}{\shortstack[l]{Partitioning Method\\+ tentative: Ensemble\\+ estimation: NES}} & \multirow{2}{*}{Random} & \multirow{2}{*}{77881} & \multirow{2}{*}{100} \\
    &&&\\
    & \textbf{Uniform} & \textbf{49797} & \textbf{100} \\  \hline \hline
    \multirow{3}{*}{\shortstack[l]{Estimation Method\\+ tentative: Ensemble\\+ partition: Uniform}} & \multirow{2}{*}{FD} & \multirow{2}{*}{61585} & \multirow{2}{*}{70} \\
    &&&\\
    &\textbf{NES} & \textbf{49797} & \textbf{100} \\
    \bottomrule
  \end{tabular}
\end{table}

\textbf{V-BAD Setting.} For tentative perturbations, we average the perturbations extracted from the three image models to obtain the final perturbations. For partitioning, we use uniform partitioning to get $8 \times 8$ patches per frame and set the NES population size of each estimation as 48, which works consistently well across different datasets in terms of both success rate and number of queries.
For search variance $\sigma$ in NES, we set it to $10^{-6}$ for the targeted attack setting and $10^{-3}$ for the untargeted attack setting. This is because a targeted attack needs to keep the target class in the top-1 class list to get the score of the targeted class, while the aim of an untargeted attack is to remove the current class from top-1 class position, which allows a large search step. We use PGD attack for step-wise perturbations with dynamically chosen step size.
For the targeted attack, we adjust the step size $\alpha$ and epsilon decay $\Delta\epsilon$ dynamically. If the ratio of failing to maintain the adversarial class is higher than a threshold of 50\%, we halve the step size $\alpha$. If we fail to reduce the perturbations size $\epsilon$ after 100 times, we halve the epsilon decay $\Delta\epsilon$.

\begin{table}[t]
\centering
  \caption{Targeted attacks on UCF-101/HMDB-51/Kinetics-400 against I3D/CNN+LSTM models. Best results are in \textbf{bold}.}
  \label{tab:targeted}
  \small
  \setlength{\tabcolsep}{0.25em} 
  \begin{tabular}{c|r|rr|rr|rr}
    \toprule
    \multirow{2}{1cm}{\centering Target Model} & \multirow{2}{*}{Attack} & \multicolumn{2}{c|}{UCF-101} & \multicolumn{2}{c|}{HMDB-51} & \multicolumn{2}{c}{Kinetics-400}\\
    & & ANQ & SR (\%) & ANQ & SR (\%) & ANQ & SR (\%)\\
    \midrule 
    \multirow{3}{*}{I3D} & SR-BAD & 67909 & 96.0 & 40824 & 96.1 & 63761 & 98.0\\
    & P-BAD & 104986 & 96.0  & 62744 & 96.8 & 84380 & 97.0 \\
    & \textbf{V-BAD} & \textbf{60687} & \textbf{98.0} & \textbf{34260} & \textbf{96.8} & \textbf{54528} & \textbf{100.0} \\
    \hline
    \multirow{3}{1cm}{\centering CNN + LSTM} & SR-BAD & 147322 & 45.5 & 67037 & 82.4 & 109314 & 73.0\\
    & P-BAD & 159723 & 60.4 & 72697 & 90.2 & 117368 & 85.0 \\
    & \textbf{V-BAD} & \textbf{84294} & \textbf{93.1} & \textbf{44944} & \textbf{98.0} & \textbf{70897} & \textbf{97.0}\\
    \bottomrule
  \end{tabular}
\end{table}

\subsection{Ablation Study of V-BAD}
In this section, we evaluate variants of V-BAD, with different types of 1) tentative perturbations, 2) partitioning methods and 3) estimation methods. Experiments were conducted on a subset of 20 randomly selected categories from UCF-101 dataset. 

\textbf{Tentative Perturbations.}
We first evaluate V-BAD with the three different types of tentative perturbations discussed in Section \ref{sec:tent}: 1) Random, 2) Static, and 3) Transferred. For transferred, we test two different strategies, with either a single image model ResNet-50 (denoted as ``Single") or an ensemble of the three considered image models (denoted as ``Ensemble"). The partitioning and estimation methods were set to Uniform and NES respectively. The results in terms of success rate (SR) and average number of queries (ANQ) can be found in Table \ref{tab:tent}. A clear improvement can be observed for the use of transferred tentative perturbations compared to static or random perturbations. The number of queries required for successful attacks was dramatically reduced from more than $10^5$ to less than $6 \times 10^4$ by using only a single image model. This confirms that an image model alone can provide effective guidance for attacking video models. The number was further reduced to around $5 \times 10^4$ via the use of an ensemble of three image models. Compared to random perturbations, static perturbations require fewer queries to succeed, with $50\%$ less queries due to the reduced exploration space. 
However, static perturbations have a much lower success rate ($70\%$) than random perturbations ($95\%$). This indicates that fixed directions can generate adversarial examples faster with restricted exploration space, however, such restrictions may cause the attack easily stuck in some local optima, without the exploration over other directions that could lead to potentially more powerful attacks.

\begin{table}[t]
\centering
  \caption{Untargeted attacks on UCF-101/HMDB-51/Kinetics-400 against I3D/CNN+LSTM models. Best results are in \textbf{bold}.}
  \label{tab:untargeted}
  \small
  \setlength{\tabcolsep}{0.25em} 
  \begin{tabular}{c|r|rr|rr|rr}
    \toprule
    \multirow{2}{1cm}{\centering Target Model} & \multirow{2}{*}{Attack} & \multicolumn{2}{c|}{UCF-101} & \multicolumn{2}{c|}{HMDB-51} & \multicolumn{2}{c}{Kinetics-400}\\
    & & ANQ & SR (\%) & ANQ & SR (\%) & ANQ & SR (\%)\\
    \midrule 
    \multirow{3}{*}{I3D} & SR-BAD & 5143 & 98.0 & 1863 & 100.0 & 1496 & 100.0\\
    & P-BAD & 11571 & 98.0  & 4162 & 100.0 & 3167 & 100.0 \\
    & \textbf{V-BAD} & \textbf{3642} & \textbf{100.0} & \textbf{1152} & \textbf{100.0} & \textbf{1012} & \textbf{100.0} \\
    \hline
    \multirow{3}{1cm}{\centering CNN + LSTM} & SR-BAD & 8674 & 100.0 & 684 & 100.0 & 1181 & 100.0\\
    & P-BAD & 12628 & 100.0 & 1013 & 100.0 & 1480 & 100.0 \\
    & \textbf{V-BAD} & \textbf{784} & \textbf{100.0} & \textbf{197} & \textbf{100.0} & \textbf{293} & \textbf{100.0}\\
    \bottomrule
  \end{tabular}
\end{table}

\textbf{Partitioning Methods.}
In this experiment, we investigate the two types of partitioning methods introduced in Section \ref{sec:group_rect}, namely, Random and Uniform. For tentative perturbations, we use the best method found in the previous experiments - Ensemble. Results are reported in Table \ref{tab:tent}. As can be seen, both partitioning methods can find successful attacks, but the use of uniform partitioning significantly reduces the number of queries by $36\%$, to around $5 \times 10^4$ from $8 \times 10^4$ of random partitioning. This is because a tentative perturbations generated from image models often contains certain local patterns, but random partitioning tends to destroy such locality. Recall that partitioning is applied in every step of perturbations, and as such, uniform partitioning can help to maintain stable and consistent patches across different iteration steps. This allows the rectification to make continuous corrections to the same local patches.

\textbf{Estimation Methods.}
As we mentioned in Section \ref{sec:group_rect}, any derivative-free (or black-box) optimization methods can be used to estimate the rectification weights. Here, we compare two of the methods that have been used for black-box image attacks: FD (Finite Difference) \cite{bhagoji2018practical} and NES (Natural Evolution Strategies) \cite{ilyas2018black}. For a fair comparison, we made some adjustments to the number of patches used by the FD estimator, so that FD and NES require a similar number of queries per update. The results are reported in Table \ref{tab:tent}. NES demonstrates a clear advantage over FD: FD only achieves 70\% success rate within the query limit (\eg $3 \times 10^5$), while NES has a 100\% success rate. And the number of queries required by successful NES attacks is roughly $20\%$ less than that of FD attacks.

Based on the ablation results, in the following experiments, we set the V-BAD to: ensemble tentative perturbations, uniform partitioning and NES rectification estimator.

\begin{figure*}
    \centering
    \subcaptionbox{Results on UCF-101\label{img:as_gamma}}{\includegraphics[width=0.33\linewidth]{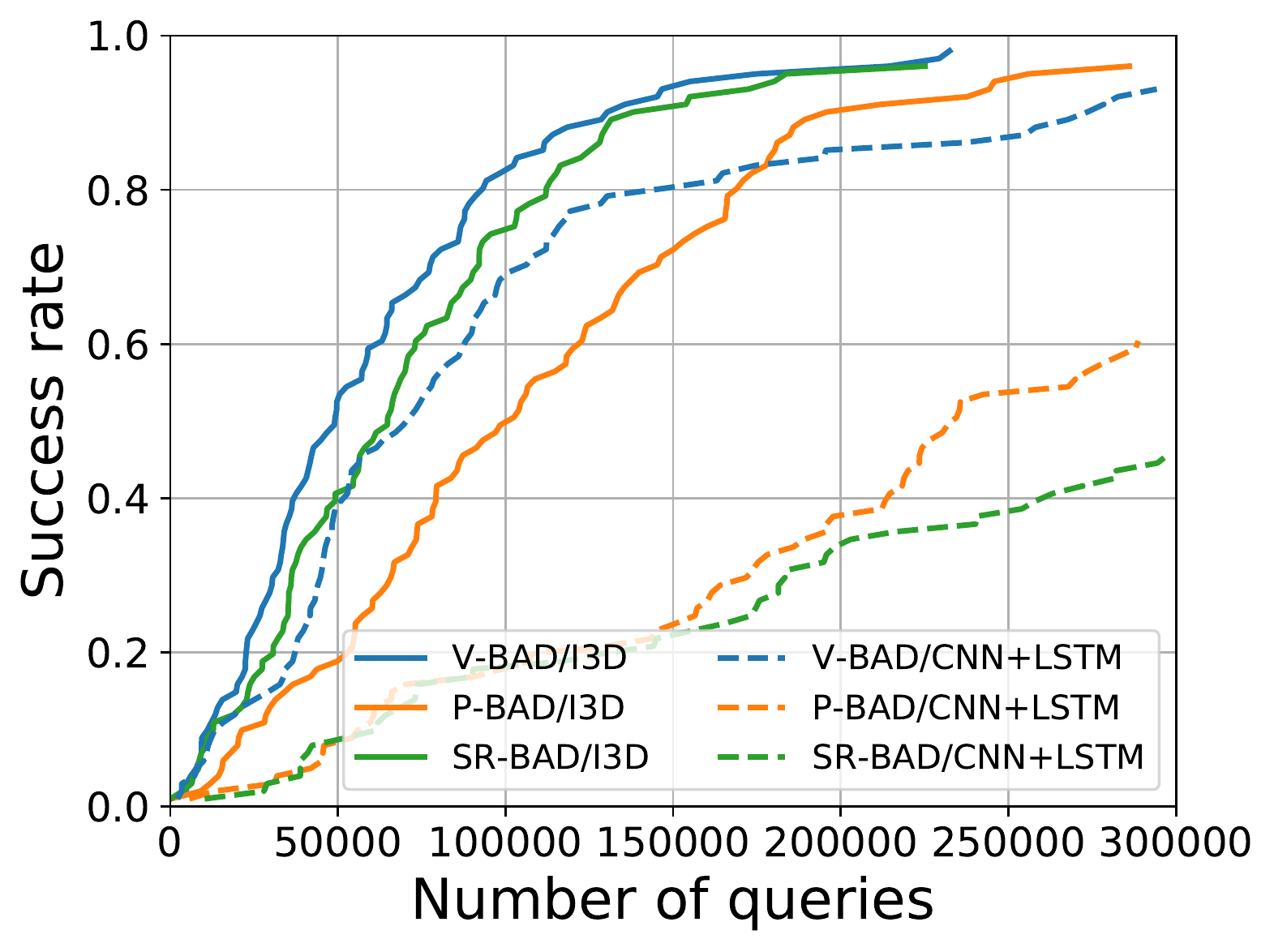}}%
    \hfill
    \subcaptionbox{Results on HMDB-51}{\includegraphics[width=0.333\linewidth]{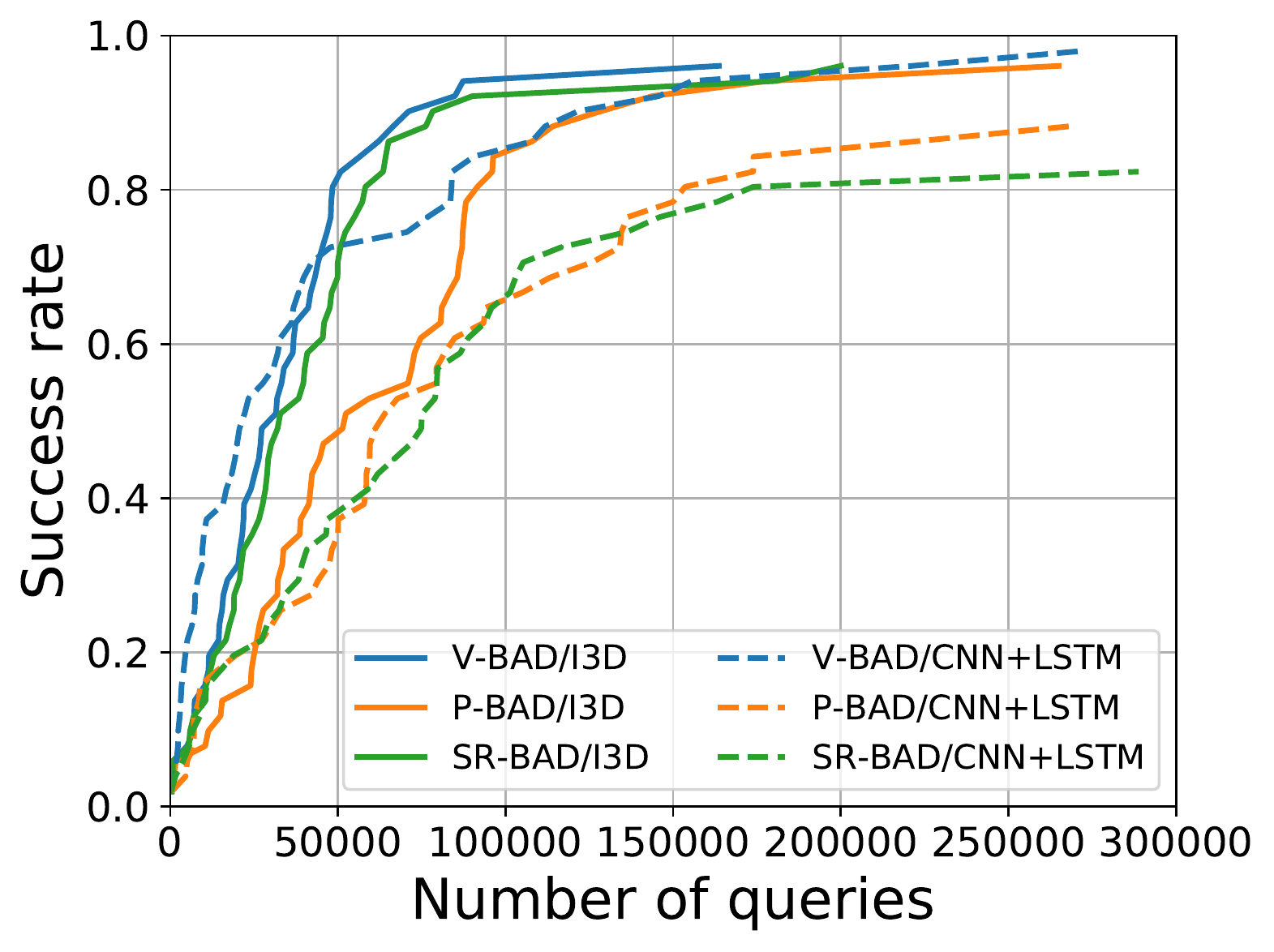}}%
    \hfill
    \subcaptionbox{Results on Kinetics-400}{\includegraphics[width=0.333\linewidth]{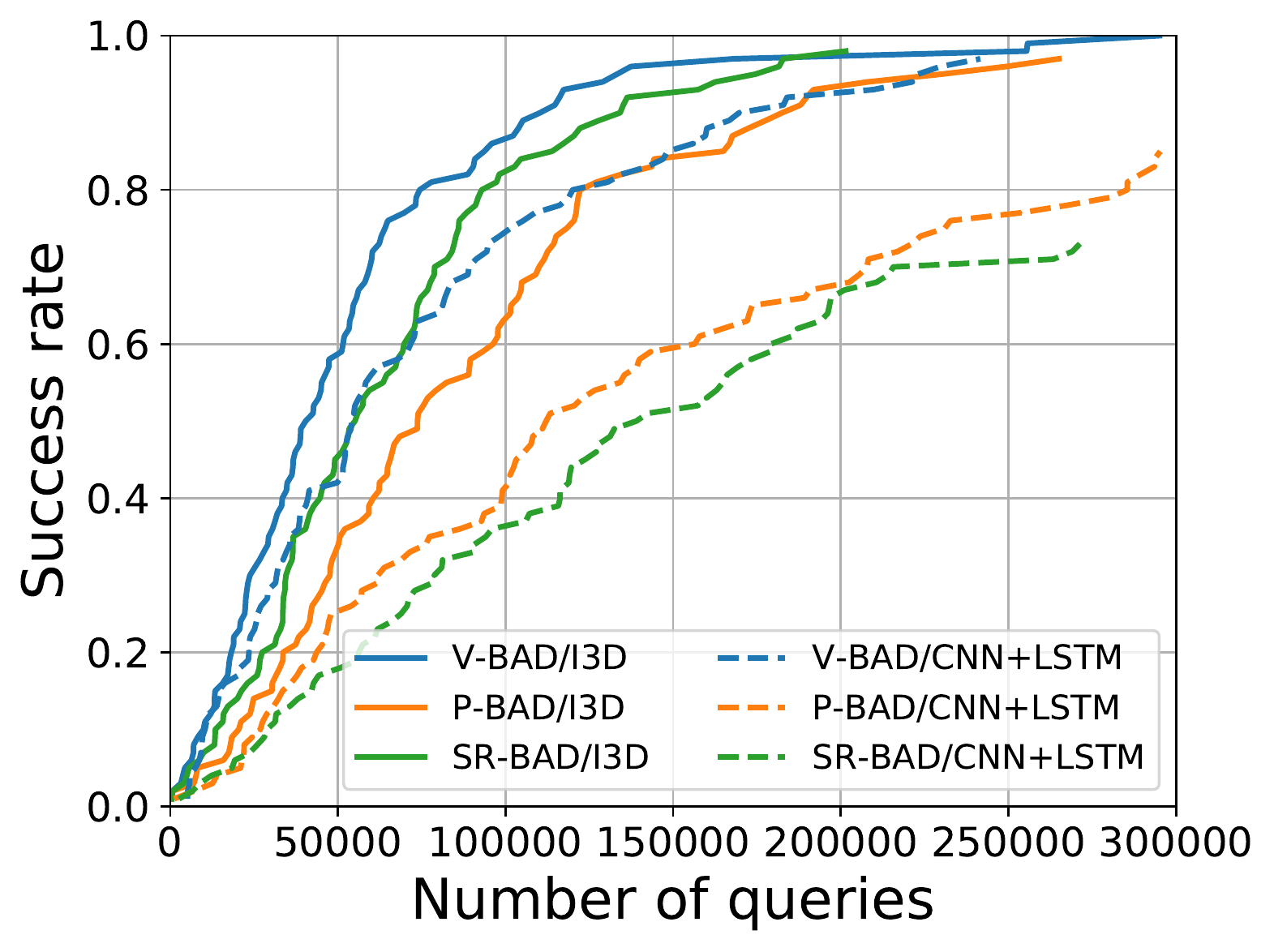}}%
    \caption{Comparative results for targeted attack.}
    \label{fig:exp}
\end{figure*}

\subsection{Comparison to Existing Attacks}
In this section, we compare our V-BAD framework with two existing state-of-the-art black-box image attack methods \cite{bhagoji2018practical,ilyas2018black}. Instead of directly applying the two image attack methods to videos, we instead incorporate their logic into our V-BAD framework to obtain two variants of V-BAD.

\textbf{Baselines.} The first baseline method is the pixel-wise adversarial gradient estimation using NES, proposed in \cite{ilyas2018black}. This can be easily achieved by using static tentative perturbations and setting the patches in V-BAD to pixels, \ie, each pixel is a patch. We denote this baseline variant of V-BAD by P-BAD. 
The NES population size for P-BAD is set as 96, since there are many more parameters to estimate. The second baseline method is grouping-based adversarial gradient estimation using FD, proposed by \cite{bhagoji2018practical}. This method explores random partitioning to reduce the large number of queries required by FD. Accordingly, we use the variant of V-BAD that utilizes static tentative perturbations and random partitioning, and denote it by SR-BAD. Different from its original setting with FD, here we use NES for SR-BAD which was found more efficient in our previous experiments.
It is worth mentioning that a dimensionality reduction technique (\eg PCA) was also explored in \cite{bhagoji2018practical} to project inputs to low dimensional features. Although it increased attack success rate on simple black-white images (\eg MNIST \cite{lecun1990handwritten}), it did not help attacks on natural images (\eg CIFAR10 \cite{krizhevsky2009learning}), due to the information loss caused by the projection.

\textbf{Targeted Black-box Video Attack.}
Comparison results for targeted attacks are reported in Table \ref{tab:targeted}. Among the three methods, V-BAD achieves the best success rates, consistently using least number of queries across the three datasets and two recognition models. Specifically, V-BAD only takes $(3.4 \sim 8.4) \times 10^4$ queries to achieve a success rate of above $93\%$. Note that this is comparable to state-of-the-art black-box image attacks \cite{ilyas2018black}. Comparing P-BAD and V-BAD, pixel-wise estimation by P-BAD does not seem to yield more effective attacks, whereas on the contrary, patition-based rectifications by V-BAD not only reduces $\sim 50\%$ queries, but also leads to more successful attacks. Compare the performance on different target models, an obvious degradation of performance can be observed on CNN+LSTM model. This is because CNN+SLTM has a lower accuracy than I3D, making it relatively robust to targeted attacks (not for untargeted attacks), an observation that is consistent with findings in \cite{tsipras2018robustness}. However, this impact is the least significant on V-BAD where the accuracy decreases less than 5\%, while the accuracy of P-BAD and SR-BAD has a huge drop, especially on UCF-101(from 96.0\% to 45.5\%). This is further illustrated in Figure \ref{fig:exp}, showing change in success rate with the number of queries. This can probably be explained by the better transferability of transferred tentative perturbations on CNN+LSTM than I3D due to the similar 2D CNN used in CNN+LSTM video model. As in Figure \ref{fig:exp}, the advantage of better transferability even overcomes the low accuracy drawback of CNN+LSTM on HMDB-51 dataset: V-BAD/CNN+LSTM is above V-BAD/I3D for the first $4 \times 10^4$ queries.
A targeted video adversarial examples generated by V-BAD is illustrated in Figure \ref{fig:advimgs}, where video on the top is the original video with the correct class and video at the bottom is the video adversarial example misclassified as the adversarial class.

\begin{table}[t]
\centering
  \caption{Cosine similarity between varies tentative or rectified gradients and the actual gradient.}
  \small
  \setlength{\tabcolsep}{0.4em} 
  \begin{tabular}{cccc}
    \toprule
    Tentative & Static & Random & Transferred  \\
    \hline
    Cosine & $7.177\times 10^{-5}$ & $-1.821\times 10^{-5}$ & $-2.743\times 10^{-4}$ \\
    \hline
    Rectified & SR-BAD & P-BAD & V-BAD \\
    \hline
    Cosine & $3.480\times 10^{-3}$& $3.029\times 10^{-3}$ & $4.661\times 10^{-3}$ \\
    \bottomrule
  \end{tabular}
  \label{tab:cosine}
\end{table}

\textbf{Untargeted Black-box Video Attack.}
Results for untargeted attacks are reported in Table \ref{tab:untargeted}. Compared to targeted attacks, untargeted attacks are much easier to achieve, requiring only $\sim 10\%$ queries of targeted attacks. Compared to other baselines, V-BAD is the most effective and efficient attack across all datasets and recognition models. It only takes a few hundred queries for V-BAD to completely break the CNN+LSTM model. 

Both attacks indicate that video models are as vulnerable as image models to black-box adversarial attacks. This has serious implications for the video recognition community to consider.

\subsection{Gradient Estimate Quality}
We further explore the quality of various tentative perturbations and rectified perturbations generated by different variants of V-BAD. We measure the perturbations quality by calculating the cosine similarity between the ground-truth adversarial gradient and the tentative/rectified perturbations. The results are based on 20 random runs of the attacks on 50 videos randomly chosen from UCF-101, and are reported in Table \ref{tab:cosine}. Consistent with the comparison experiments, V-BAD generates the best gradient estimates and P-BAD has the worst estimation quality. All the rectified perturbations are much better than the tentative perturbations. This verifies that tentative perturbations can be significantly improved by proper rectification.
One interesting observation is that the transferred tentative perturbations (from an ensemble of ImageNet models) have a large negative cosine similarity, which is opposite to our design. One explanation could be that there is a huge gap between the image model and the video model. However, note that while the transferred perturbations is opposite to the gradient, it serves as a good initialization and yields better gradient estimation after rectification. It is noteworthy that there is still a considerable gap between the gradient estimate and the actual gradient. From one angle, it reflects that we do not need very accurate gradient estimation to generate adversarial examples. From another angle, it suggests that black-box attack based on gradient estimation has great scope for further improvement.

\section{Conclusion}
In this paper, we investigated the problem of black-box adversarial attack against video recognition models, and proposed the first framework, V-BAD, for the generation of video adversarial examples through only black-box queries to a video model. To address efficiency issues caused by the high dimensionality of videos, we decoupled adversarial gradient estimation into a two-step process: tentative perturbations transfer followed by partition-based rectification with NES estimation method. We demonstrated the effectiveness and efficiency of V-BAD by attacking two state-of-the-art video recognition models on three benchmark video datasets. Compared to existing black-box methods, V-BAD achieved high success rate using significantly less queries for both targeted and untargeted attacks. Our results suggest that video models are also highly vulnerable to black-box adversarial attacks, and that effective defenses for video models will need to be developed for secure video recognition.

\bibliographystyle{IEEEtran}
\bibliography{mm2019}

\end{document}

%% file: pseudocode/framework.tex
\begin{algorithm}[t]
   \caption{Targeted V-BAD attack}
   \label{alg:framework}
\begin{algorithmic}
    \STATE {\bfseries Input:} Top-1 probability $P(y|\xx)$ with respect to classifier $f$, target class $\ya$ and video $\xx$
    \STATE {\bfseries Output:}  Adversarial video $\xa$ with
    $||\xa - \xx||_\infty \leq \epsilon$
    \STATE {\bfseries Parameters:} Perturbation bound $\epsilon_{adv}$, epsilon decay $\Delta\epsilon$, PGD step size $\alpha$
    \STATE $\epsilon \gets 1$
    \STATE $\xa \gets$ video of target class $\ya$
    \WHILE{$\epsilon > \epsilon_{adv}$}
    \STATE $\vv = \mathbf{0}$
    \STATE $\hh = \phi(\xa)$
    \STATE $\bU = G(\hh)$
    \STATE $\hat{\vv} = \vv + \nabla_{\vv}\la(\xa+R(\vv, \bU))$
    \STATE $\hat{\gg} = R(\hat{\vv}, \bU)$
    
    
    \STATE $\hat{\epsilon} \gets \epsilon - \Delta\epsilon$
    
    \STATE $\hat{\xx}_{adv} \gets \textsc{Clip}(\xa - \alpha\cdot \hat{\gg}, \xx - \hat{\epsilon}, \xx + \hat{\epsilon})$
    
        \IF{$\ya = \textsc{Top-1}(P(\cdot|\hat{\xx}_{adv}))$}
        \STATE $\xa \gets \hat{\xx}_{adv}$
        \STATE $\epsilon \gets \hat{\epsilon}$
    	\ELSE
        \STATE $\hat{\xx}_{adv} \gets \textsc{Clip}(\xa - \alpha\cdot \hat{\gg}, \xx - \epsilon, \xx + \epsilon)$
    	    \IF{$\ya = \textsc{Top-1}(P(\cdot|\hat{\xx}_{adv}))$}
    	        \STATE $\xa \gets \hat{\xx}_{adv}$
            \ENDIF
    	\ENDIF
    \ENDWHILE
    \STATE {\bfseries return} $\xa$
\end{algorithmic}
\end{algorithm}

%% file: pseudocode/rectify.tex
\begin{algorithm}[t]
   \caption{NES Estimation of Patch Rectification}
   \label{alg:nes}
\begin{algorithmic}
    \STATE {\bfseries Input:} Adversarial loss $\la(\xx)$ with respect to input $\xx$ and tentative perturbation patches $\bU$
    \STATE {\bfseries Output:} Estimate of $\nabla_{\vv}\la(\xx+R(\vv, \bU))$
    \STATE {\bfseries Parameters:} Search variance $\sigma$, sampling size $\lambda$
   \STATE $\hat{\vv} \gets \bm{0}_M$
   
    
   \FOR{$k=1$ {\bfseries to} $\lambda/2$}
	\STATE $ \bm\delta_{k} \gets \mathcal{N}(\bm{0}_M, \bm{I}_{M\times M})$
    \STATE $\zz_{2k-1} \gets \la(\xx + R(\vv + \sigma\bm\delta_{k}, \bU))$
	
	\STATE $\hat{\vv} \gets \hat{\vv} + \bm\delta_{k} \cdot \text{TransformedAdvLoss}(\zz_{2k-1})$
	
	\STATE $\zz_{2k} \gets \la(\xx - R(\vv + \sigma\bm\delta_{k}, \bU))$
	\STATE $\hat{\vv} \gets \hat{\vv} - \bm\delta_{k} \cdot \text{TransformedAdvLoss}(\zz_{2k})$
   \ENDFOR
   \STATE $\hat{\vv} \gets \frac{1}{\lambda\sigma} \hat{\vv}$
   \STATE {\bfseries return} $\hat{\vv}$
\end{algorithmic}
\end{algorithm}